\documentclass[lettersize,journal]{IEEEtran}
\usepackage{graphicx}%
\usepackage{multirow}%
\usepackage{amsmath,amssymb,amsfonts}%
\usepackage{amsthm}%
\usepackage{mathrsfs}%
\usepackage[dvipsnames]{xcolor}
\usepackage{textcomp}%
\usepackage{manyfoot}%
\usepackage{booktabs}%
\usepackage{algorithm}%
\usepackage{algorithmicx}%
\usepackage{algpseudocode}%
\usepackage{listings}%
\usepackage{xcolor}
\usepackage{tabularx}
\usepackage{makecell}
\usepackage{supertabular}
\usepackage{float}
\usepackage{threeparttable}
\usepackage{hyperref}
\usepackage{siunitx}
\usepackage{booktabs}
\usepackage[backend=biber,style=ieee,bibencoding=utf8,hyperref=true]{biblatex}
\usepackage{color, colortbl}
\usepackage{subcaption} 

\DeclareUnicodeCharacter{0301}{\'{e}}
\def\BibTeX{{\rm B\kern-.05em{\sc i\kern-.025em b}\kern-.08em
    T\kern-.1667em\lower.7ex\hbox{E}\kern-.125emX}}
\begin{document}



\global\long\def\dq#1{\underline{\boldsymbol{#1}}}%

\global\long\def\quat#1{\boldsymbol{#1}}%

\global\long\def\mymatrix#1{\boldsymbol{#1}}%

\global\long\def\myvec#1{\boldsymbol{#1}}%

\global\long\def\mapvec#1{\boldsymbol{#1}}%

\global\long\def\dualvector#1{\underline{\boldsymbol{#1}}}%

\global\long\def\dual{\varepsilon}%

\global\long\def\dotproduct#1{\langle#1\rangle}%

\global\long\def\norm#1{\left\Vert #1\right\Vert }%

\global\long\def\mydual#1{\underline{#1}}%

\global\long\def\hamilton#1#2{\overset{#1}{\operatorname{\mymatrix H}}\left(#2\right)}%

\global\long\def\hamiquat#1#2{\overset{#1}{\operatorname{\mymatrix H}}_{4}\left(#2\right)}%

\global\long\def\hami#1{\overset{#1}{\operatorname{\mymatrix H}}}%

\global\long\def\tplus{\dq{{\cal T}}}%

\global\long\def\getp#1{\operatorname{\mathcal{P}}\left(#1\right)}%

\global\long\def\getd#1{\operatorname{\mathcal{D}}\left(#1\right)}%

\global\long\def\swap#1{\text{swap}\{#1\}}%

\global\long\def\imi{\hat{\imath}}%

\global\long\def\imj{\hat{\jmath}}%

\global\long\def\imk{\hat{k}}%

\global\long\def\real#1{\operatorname{\mathrm{Re}}\left(#1\right)}%

\global\long\def\imag#1{\operatorname{\mathrm{Im}}\left(#1\right)}%

\global\long\def\imvec{\boldsymbol{\imath}}%

\global\long\def\vector{\operatorname{vec}}%

\global\long\def\mathpzc#1{\fontmathpzc{#1}}%

\global\long\def\cost#1#2{\underset{\text{#2}}{\operatorname{\text{cost}}}\left(\ensuremath{#1}\right)}%

\global\long\def\diag#1{\operatorname{diag}\left(#1\right)}%

\global\long\def\frame#1{\mathcal{F}_{#1}}%

\global\long\def\ad#1#2{\text{Ad}\left(#1\right)#2}%



\global\long\def\argminimone#1#2#3#4{\begin{aligned}#1\:  &  \underset{#2}{\arg\!\min}  &   &  #3\\
  &  \text{subject to}  &   &  #4 
\end{aligned}
 }%

\global\long\def\minimtwo#1#2#3#4{ \begin{aligned} &  \underset{#1}{\min}  &   &  #2 \\
  &  \text{subject to}  &   &  #3 \\
  &   &   &  #4 
\end{aligned}
 }%

\global\long\def\minimone#1#2#3{ \begin{aligned} &  \underset{#1}{\min}  &   &  #2 \\
  &  \text{subject to}  &   &  #3 
\end{aligned}
 }%

\global\long\def\argminimtwo#1#2#3#4#5{ \begin{aligned}#1\:  &  \underset{#2}{\arg\!\min}  &   &  #3 \\
  &  \text{subject to}  &   &  #4\\
  &   &   &  #5 
\end{aligned}
 }%

\global\long\def\argmaximtwo#1#2#3#4#5{ \begin{aligned}#1\:  &  \underset{#2}{\arg\!\max}  &   &  #3 \\
  &  \text{subject to}  &   &  #4 \\
  &   &   &  #5 
\end{aligned}
 }%

\title{Automated Robotic Needle Puncture for Percutaneous Dilatational Tracheostomy}
\author{Yuan Tang, Bruno V. Adorno,~\IEEEmembership{Senior Member,~IEEE,} Brendan A. McGrath and Andrew Weightman
\thanks{(Corresponding author: Andrew Weightman)}
\thanks{Y. Tang and A. Weightman are with the Manchester Centre for Robotics and AI and the Department of Mechanical and Aerospace Engineering, University of Manchester, Manchester, UK (yuan.tang@manchester.ac.uk, andrew.weightman@manchester.ac.uk)}
\thanks{B. V. Adorno is with the Manchester Centre for Robotics and AI and the Department of Electrical and Electronic Engineering, University of Manchester, Manchester, UK (bruno.adorno@manchester.ac.uk)}
\thanks{B. A. McGrath is with Manchester University Foundation Trust Wythenshawe Hospital Intensive Care Unit, Manchester, UK (brendan.mcgrath@mft.nhs.uk)}
}

\maketitle

\begin{abstract}

Percutaneous dilatational tracheostomy (PDT) is frequently performed on patients in intensive care units for prolonged mechanical ventilation. The needle puncture, as the most critical step of PDT, could lead to adverse consequences such as major bleeding and posterior tracheal wall perforation if performed inaccurately. Current practices of PDT puncture are all performed manually with no navigation assistance, which leads to large position and angular errors ($\geq 5$ mm and $30^\circ$). To improve the accuracy and reduce the difficulty of the PDT procedure, we propose a system that automates the needle insertion using a velocity-controlled robotic manipulator. Guided using pose data from two electromagnetic sensors, one at the needle tip and the other inside the trachea, the robotic system uses an adaptive constrained controller to adapt the uncertain kinematic parameters online and avoid collisions with the patient's body and tissues near the target. Simulations were performed to validate the controller's implementation, and then four hundred PDT punctures were performed on a mannequin to evaluate the position and angular accuracy. The absolute median puncture position error was $1.7$ mm (IQR: $1.9$ mm) and midline deviation was $4.13^{\circ}$ (IQR: $4.55^{\circ}$), measured by the sensor inside the trachea. The small deviations from the nominal puncture in a simulated experimental setup and formal guarantees of collision-free insertions suggest the feasibility of the robotic PDT puncture.

\end{abstract}

\begin{IEEEkeywords}
Percutaneous dilatational tracheostomy, Surgical robots, Constrained control, Adaptive control
\end{IEEEkeywords}

\section{Introduction}

\IEEEPARstart{P}{ercutaneous} dilatation tracheostomy (PDT) is commonly performed on patients managed in intensive care units (ICUs) \cite{durb2010,de2007}. The most common indication is to facilitate the gradual reduction of mechanical ventilatory support and reduction/cessation of sedation \cite{cheu2014}. With comparable complication rates but shorter hospital stays and the ability to be performed at the ICU bedside, PDT becomes an appealing alternative to the traditional surgical approach \cite{al2005}.

\begin{figure}[t]
\centerline{\includegraphics[width=\columnwidth]{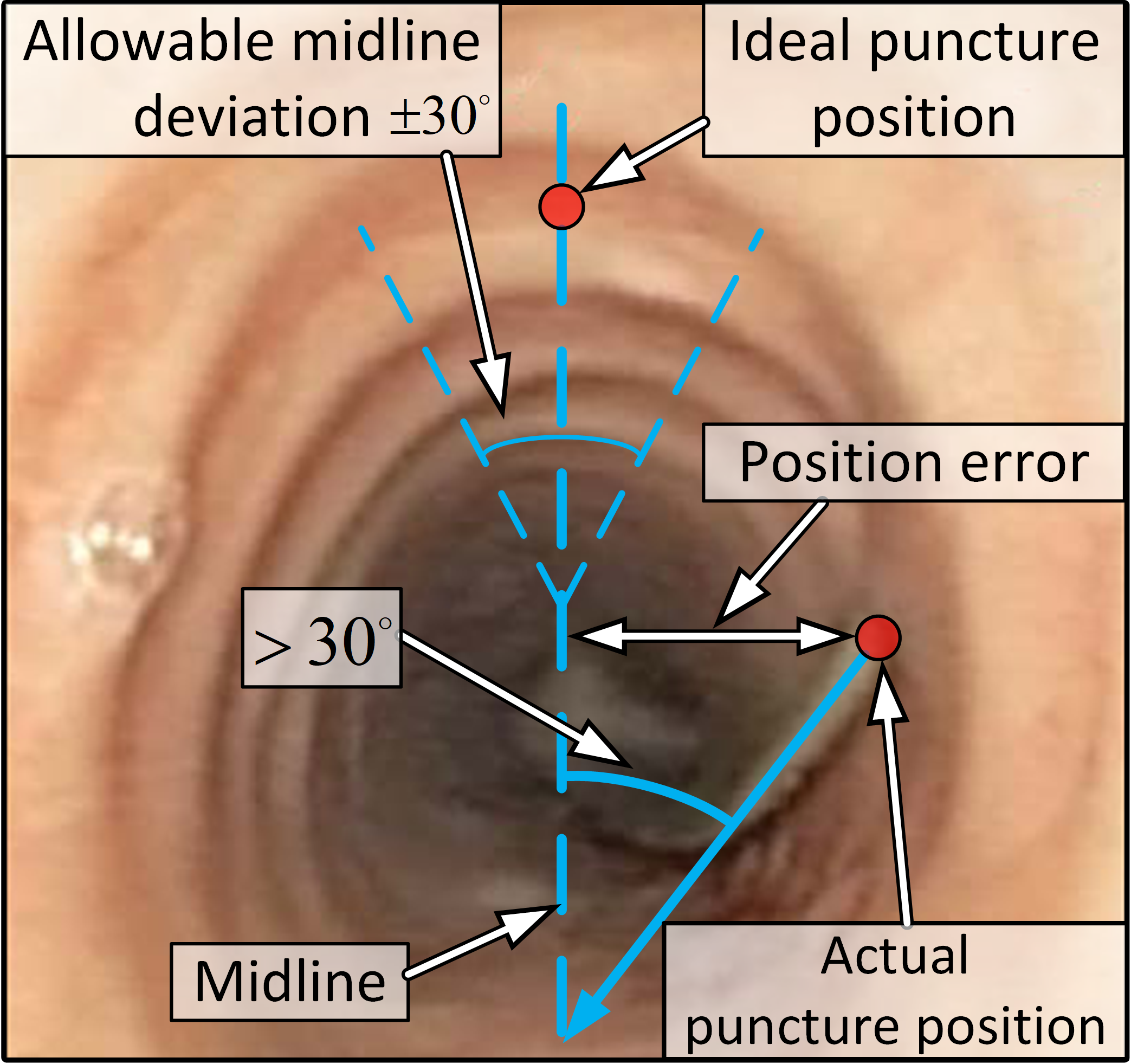}}
\caption{Bronchoscopic view of an extremely lateral PDT puncture \cite{ruda2012}. Lateral punctures could lead to complications such as tracheostomy tube misplacements and posterior tracheal wall damage.}
\label{midline}
\end{figure}

The main steps of PDT include a needle puncture followed by dilation to expand the incision. Among all steps, the quality of the needle puncture dominates the outcome of the whole procedure. Without the exposure of the trachea (below the vocal cords, lying at the middle of the anterior neck), the incision position is determined by palpating the patient's neck \cite{durb2010} with visual assistance provided by additional equipment such as a bronchoscope or ultrasound so that the clinician can visualize whether the needle is being inserted correctly. However, both techniques cannot accurately guide the needle movement during puncture due to limitations of a 2D visualization of motion that happens intrinsically in a 3D space, such as a lack of accurate depth perception. Ideally, the puncture should be performed at the midline perpendicular to the posterior tracheal wall (12 o'clock direction) seen from the cross-section, as illustrated in Fig.~\ref{midline}. Furthermore, the needle must puncture through the anterior tracheal wall between the first and second (or second and third) tracheal rings, regarding the lower limit of the inner diameter of a normal adult trachea to be 10--13 mm \cite{brea1984}. Inaccurate or lateral needle punctures occur during the PDT needle puncture due to (1) a lack of navigation assistance in current visual techniques, and (2) the impracticality of the palpation-only method for patients with thick pretracheal tissues. Previous studies quantified that the allowable angular error of needle puncture is $\pm30^{\circ}$ of midline deviation \cite{ruda2014}. Considering the nature of PDT, the position error smaller than the radius of trachea and the midline deviation angle smaller than $30^{\circ}$ are considered clinically valid. 

Some PDT complications are related to human errors made during the procedure due to the lack of guidance. Inaccurate or lateral punctures may lead to major bleeding, tracheostomy tube misplacement or reinsertion, and damage to the posterior tracheal wall or thyroid caused by the needle \cite{ruda2014,zouk2021}. Ideally, the puncture should be accurately performed on the first attempt. However, even experienced operators cannot guarantee accurate needle placements in the first puncture due to a lack of feedback on the position and direction of the needle with respect to the desired puncture pose inside the trachea.

Robotic technologies have been widely adopted in various types of minimally invasive surgery (MIS), such as neurosurgery \cite{hu2015} and joint arthroplasty \cite{wolf2005}. Surgical robots have demonstrated the potential to replace some manual procedures in MIS that require a high level of precision \cite{corc2005}. Currently, few prototypes have been developed to perform PDT, and all perform the puncture from inside the trachea \cite{boty2018,xiao2020}. Given the clinical preference and the lower difficulty of the traditional outside-in technique \cite{veen2008}, robotic systems that guide and automate the needle puncture using the outside-in approach could be developed to achieve high precision, release the operator from this step, and decrease the procedural difficulty. Since needle puncture is the most dangerous and critical step of PDT, robots may have the potential to improve accuracy, increase the success rate of placing the needle accurately in the first attempt, or even perform punctures autonomously. 

This research proposes a robotic system that replaces manual needle insertion to improve puncture precision and reduce the difficulty of the PDT procedure. An electromagnetic sensing system composed of one transmitter placed near the patient's head, one sensor for the acquisition of target puncture position and direction, and one sensor for needle navigation is used to guide a robot manipulator with a needle at its end-effector. Since PDT puncture requires accurate needle movements, avoidance of any robot-patient collisions (a main requirement for surgical robots \cite{diaz2017}), adaptation to uncertain nonlinear kinematic parameters, and preferably an automated process, we use the adaptive constrained controller \cite{mari2022} based on vector-field inequalities (VFIs) \cite{mari2018,mari2019} as it fulfils all the aforementioned requirements. In this framework, the dynamics between geometric primitives representing the obstacles in the workspace and the robot are given by differential inequalities in a constrained optimization problem. The VFIs are suitable for generating a safe trajectory for the robot end-effector while avoiding nearby constraints. Furthermore, the adaptive element of the control law compensates for the uncertain displacement between the electromagnetic transmitter and the robot, as well as the uncalibrated needle mechanical attachment to the robot, to drive the needle toward the target. 

\section{Sensor-guided adaptive constrained control for needle insertion\label{Sensor-Guided-Constrained-Adapti}}

\begin{figure}[ht]
\centerline{\includegraphics[width=1\columnwidth]{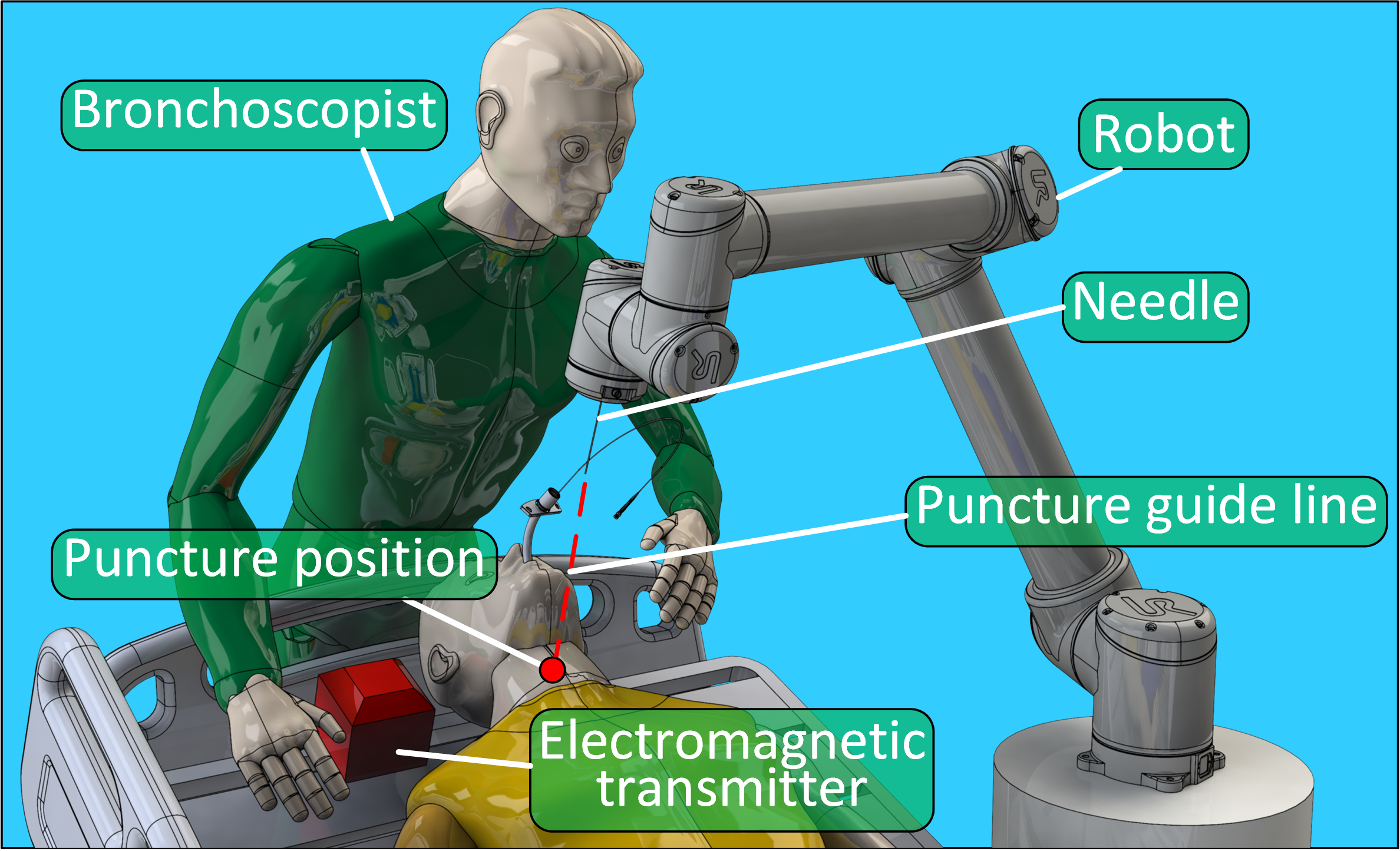}}
\caption{System hardware for automated needle puncture: a robot manipulator, an electromagnetic transmitter, and two sensors (one at the needle tip for navigating the robot, one inserted through the bronchoscope channel and reaching its tip for measuring the desired target puncture position and direction).}
\label{RobotPuncture}
\end{figure}

The system, shown in Fig.~\ref{RobotPuncture}, consists of an electromagnetic tracking system and a seven-degree-of-freedom (DoF) serial robot. The electromagnetic tracking system has two sensors that measure 3-DoF position and 3-DoF orientation data. The bronchoscope sensor is attached to the bronchoscope tip and manually inserted into the trachea to identify the target puncture position and direction. The needle sensor is attached to the needle tip for navigating the robot's movement. The robot is controlled through joint velocities, with the sensorized needle installed on its end-effector to perform the puncture automatically.

The robotic puncture is divided into three steps:
\begin{itemize}
\item \textbf{Step 1}: The robot aligns the needle with the puncture guide line.
\item \textbf{Step 2}: The robot moves to the user-defined start position above the patient’s neck without colliding with the patient's body. The robot waits for the adapting parameters to achieve a steady state before starting the next step.
\item \textbf{Step 3}: The robot advances the needle along the puncture guide line until its tip reaches the target.
\end{itemize}

\subsection{VFI-based constrained control law}

The control law consists of three main components: a constrained control law that generates the control inputs based on the nominal robot kinematic model; an adaptive law that adapts uncertain parameters of the nominal robot kinematic model to improve the estimate of the task vector (for example, the position and direction of the needle tip); and VFIs to prevent collisions with obstacles, which is achieved by defining appropriate differential inequalities directly in the control law.

\subsubsection{Nominal constrained control law}

A nominal constrained control law based on VFIs \cite{mari2019} was used to generate the velocity control inputs $\myvec u\triangleq\dot{\myvec q}\in\mathbb{R}^{n}$ for an $n$-DoF manipulator while accounting for geometric constraints in the control law explicitly. The desired task vector (e.g., the desired value for the needle tip position and direction) for the PDT puncture was given by $\boldsymbol{x}_{d}(t)\in\mathbb{R}^{m}$ and the task error was given by $\tilde{\boldsymbol{x}}(t)\triangleq\boldsymbol{x}(t)-\boldsymbol{x}_{d}(t)$, where $\myvec x(t)$ is the task vector calculated using forward kinematics. A \emph{desired} closed-loop error dynamics was defined as $\dot{\tilde{\boldsymbol{x}}}+\eta\tilde{\boldsymbol{x}}=\myvec 0$, which implies $\dot{\boldsymbol{x}}(t)-\dot{\boldsymbol{x}}_{d}(t)+\eta\tilde{\myvec x}=\myvec 0$. Using the task Jacobian $\boldsymbol{J}\in{\mathbb{R}^{m\times n}}$ satisfying $\dot{\myvec x}(t)=\myvec J(\myvec q)\dot{\myvec q}$, the control input $\boldsymbol{u}$ was given by

\begin{equation}
\argminimone{\myvec u\in}{\dot{\myvec q}}{\norm{\mymatrix J\dot{\myvec q}+\eta\tilde{\myvec x}-\dot{\myvec x}_{d}}_{2}^{2}+\lambda^{2}\norm{\dot{\myvec q}}_{2}^{2}}{\begin{array}{c}
\mymatrix B(\myvec q)\dot{\myvec q}\preceq\myvec b(\myvec q)\end{array},}
\label{quadratic}
\end{equation}

\noindent where $\eta\in(0,\infty)$ is a proportional gain determining the convergence rate and $\lambda\in\mathbb{R}$ is the damping factor to penalise large joint velocities and guarantee smoothness in the control input. The $\ell$ (scalar) linear constraints in the control inputs are represented by $\mymatrix B(\myvec q)\in{\mathbb{R}^{\ell\times n}}$ and $\myvec b(\myvec q)\in{\mathbb{R}^{\ell}}$. The $\eta$ was empirically chosen as 0.2-0.5 in different steps in the procedure so that the robot moved with different velocities while maintaining an adequate convergence rate. 

The VFIs consist of differential inequalities used to prevent collisions between the robot and geometrical primitives \cite{mari2019}. A differentiable signed distance $d\left(\myvec q\left(t\right)\right)\in\mathbb{R}$ between two collidable entities (e.g., a robot link and the patient) was defined, such that $\dot{d}\left(\myvec q\left(t\right)\right)=\mymatrix J_{d}\left(\myvec q\right)\dot{\myvec q}$, with $\boldsymbol{J}_{d}\triangleq\partial d(\myvec q)/\partial\myvec q$ being the distance Jacobian. Then, given a constant safe distance $d_{s}\in[0,\infty)$ and a signed error distance $\tilde{d}(t)\triangleq d(t)-{d_{s}}$, enforcing the inequality

\begin{equation}
\dot{\tilde{d}}(t)+\eta_{d}\tilde{d}(t)\geqslant0\iff-\boldsymbol{J}_{d}\dot{\myvec q}\leqslant\eta_{d}\tilde{d}(t)
\label{VFI-outside}
\end{equation}

\noindent ensures that $\tilde{d}(t)\geq\tilde{d}(0)e^{-\eta_{d}t}$ for all $t\geq0$ \cite{mari2019} when $\eta_{d}\in(0,\infty)$. Therefore, if $d(0)\geq d_{s}$, then $d(t)\geq d_{s}$ for all $t>0.$ An analogous reasoning applies when the inequality is reversed, in which the goal is to ensure $d(t)\leq d_{s}$ that a given primitive remains inside a safe zone.

\subsubsection{Adaptive control law\label{Adaptive-constrained-control}}

As shown in Fig.~\ref{RobotPuncture}, the robotic puncture system relies on an electromagnetic transmitter placed near the patient's head ($\leq200$ mm distance) so that the transmitter range can cover the entire procedural area. In this implementation, the transmitter frame is defined as the global reference frame. Therefore, a rigid transformation between the robot base frame and the transmitter frame is needed to map the sensor measurements from the transmitter's coordinate system to the robot's coordinate system. This transformation is inaccurate due to uncertainties in the transmitter placement. Furthermore, the needle attachment also has uncertainties associated with the mechanical attachment. Lastly, several factors might affect the sensor accuracy during the surgical procedure, such as the sensor temperature, disturbances in the electromagnetic field, and the sensor orientation \cite{shen2008}, which cannot be fully compensated for by pre-procedural calibrations.

Therefore, an adaptive control law \cite{mari2022} was used to adapt the uncertain parameters of the nominal robot kinematic model (robot base pose with respect to the transmitter frame, and the needle tip pose with respect to the robot end-effector frame) and better estimate the needle tip position and direction online. The controller adapts the estimated robot base pose with respect to the transmitter frame and the estimated needle tip pose with respect to the robot base frame. During the initial setup, those poses are initialised with manual measurements using a ruler and a protractor, which is inaccurate but sufficient to initialise the parameters that will be further adapted using the adaptive controller. Then, they are adapted online whenever new measurements of the needle tip position and direction are obtained through the electromagnetic sensor placed at the needle tip, increasing the accuracy of the needle tip positioning and alignment.

To adapt the parameters of the nominal task-space control law \eqref{quadratic}, we rewrote the objective function as $\norm{\mymatrix J_{\hat{x},q}\dot{\myvec q}+\eta\breve{\myvec x}-\dot{\myvec x}_{d}}_{2}^{2}+\lambda^{2}\norm{\dot{\myvec q}}_{2}^{2}$, where $\mymatrix J_{\hat{x},q}\triangleq\mymatrix J_{\hat{x},q}(\myvec q,\hat{\myvec a})\in\mathbb{R}^{m\times n}$ is the estimated task Jacobian and $\breve{\boldsymbol{x}}\triangleq\hat{\boldsymbol{x}}-\boldsymbol{x}_{d}$ is the estimated task-space error, with $\hat{\boldsymbol{x}}\triangleq\hat{\boldsymbol{x}}(\boldsymbol{q},\hat{\boldsymbol{a}})\in\mathbb{R}^{m}$ and $\hat{\boldsymbol{a}}(t)\in\mathbb{R}^{p}$ being the estimated task-space vector and vector of estimated robot base and needle tip parameters, respectively. The adaptation law is given by \cite{mari2022}

\begin{equation}
\argminimone{\boldsymbol{u}_{a}\in}{\dot{\hat{\myvec a}}}{\left\Vert \mymatrix J_{\hat{y},\hat{a}}\dot{\hat{\myvec a}}+\eta_{\hat{a}}\tilde{\myvec y}\right\Vert _{2}^{2}+\lambda_{a}^{2}\left\Vert \dot{\hat{\myvec a}}\right\Vert _{2}^{2}}{\begin{split}\boldsymbol{B}_{\hat{a}}\dot{\hat{\myvec a}} & \preceq\myvec b_{\hat{a}}\\
\breve{\myvec x}^{T}\mymatrix J_{\hat{x},\hat{a}}\dot{\hat{\myvec a}} & \leq0,
\end{split}
}\label{adaptation}
\end{equation}

\noindent where $\eta_{\hat{a}}\in(0,\infty)$ is the proportional gain determining the convergence rate for the adaptation, and $\tilde{\boldsymbol{y}}\triangleq\hat{\boldsymbol{y}}-\boldsymbol{y}$ is the error between the estimated needle tip pose $\hat{\boldsymbol{y}}\triangleq\hat{\boldsymbol{y}}\left(\myvec q,\hat{\myvec a}\right)$ and the pose $\boldsymbol{y}$ measured by the sensor. The term ${\lambda_{a}^{2}}{{\left\Vert \dot{\hat{\boldsymbol{a}}}\right\Vert }_{2}^{2}}$ minimizes detrimental effects when $\mymatrix J_{\hat{y},\hat{a}}\triangleq\mymatrix J_{\hat{y},\hat{a}}\left(\myvec q,\hat{\myvec a}\right)$ is ill-conditioned and guarantees that the parameters stop updating when the estimated error $\tilde{\boldsymbol{y}}$ is $\myvec 0$ \cite{mari2022}. The constraint parameters $\mymatrix B_{\hat{a}}\triangleq\mymatrix B_{\hat{a}}(\myvec q,\hat{\myvec a})\in\mathbb{R}^{\ell\times p}$ and $\myvec b_{\hat{a}}\triangleq\myvec b_{\hat{a}}(\myvec q,\hat{\myvec a})\in\mathbb{R}^{\ell}$ are used to ensure that the $\ell$ (scalar) linear constraints from \eqref{quadratic} are enforced during the adaptation. The explicit definitions of $\mymatrix B_{\hat{a}}$ and $\myvec b_{\hat{a}}$ are given in \cite{mari2022}. The second constraint ensures that the closed-loop system under the nominal and adaptive control laws is Lyapunov stable \cite{mari2022}.

\subsubsection{Constraints}

\begin{figure}[ht]
\centerline{\includegraphics[width=1\columnwidth]{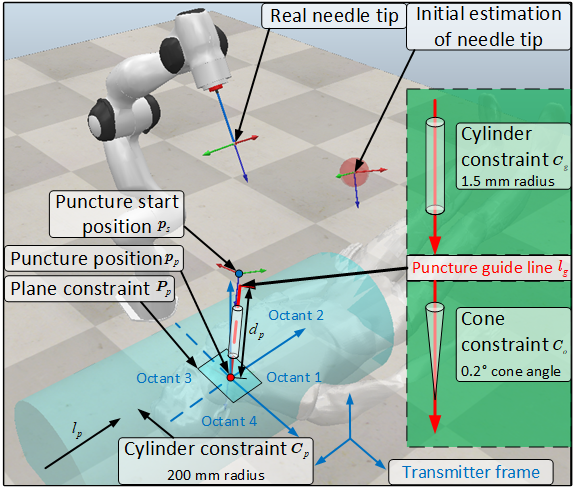}}
\caption{Simulation setup for robotic PDT needle puncture. The initial estimated needle tip position might be very inaccurate due to high uncertainty in the robot base displacement and needle attachment.}
\label{simulation}
\end{figure}

The definition of constraints given by \eqref{VFI-outside} is done by using elements of the dual quaternion algebra \cite{mari2019}, with primitives such as points, spheres, lines, cylinders, and planes already implemented in computational libraries such as DQ Robotics \cite{ador2021}. For instance, in steps 1 and 2, to ensure that an arbitrary point (expressed in pure quaternion) $\quat p_{r}(\myvec q)\in\mathbb{H}_p$ in the robot is outside the (infinite) cylinder $\myvec C_{p}=(\myvec c_{p}, R_{p})$ enclosing the patient (see Fig.~\ref{simulation}), preventing collision, where $\myvec c_{p}\in\mathcal{H}_p$ is the Plücker line (expressed in pure dual quaternion and contains information of the line direction and location \cite{seli2007}) representing the central axis and $R_{p}\in(0,\infty)$ is the radius, the corresponding VFI was defined as follows \cite{mari2019}. The cylinder's central axis was determined as $\myvec c_{p} = \myvec l_{p} + \dual\myvec m_{p}$, where $\myvec l_{p}\in\mathbb{H}^{p}$ is a pure unit quaternion representing the axis direction, and $\dual$ is the dual sign, and $\myvec m_{p}=\myvec p_{p}\times\myvec l_{p}$, with $\myvec p_{p}\in\mathbb{H}_p$ being the target puncture position (expressed in pure quaternion) located on the cylinder's axis. By defining the (squared) distance $\tilde{D}_{p_{r},C_{p}}\triangleq D_{p_{r},c_{p}}\left(\quat p_{r}(\myvec q),\myvec c_{p}\right)-R_{p}^{2}$, where $D_{p_{r},c_{p}}\left(\quat p_{r}(\myvec q),\myvec c_{p}\right)$ is the squared distance between the point $\myvec p_{r}$ and the central axis $\myvec c_{p}$, the inequality ensuring collision avoidance between $\myvec p_{r}$ and $\myvec C_{p}$ was given by

\begin{gather}
\dot{\tilde{D}}_{p_{r},C_{p}}+\eta_{c}\tilde{D}_{p_{r},C_{p}}\geq0\implies-\frac{\partial D_{p_{r},c_{p}}}{\partial\myvec q}\dot{\myvec q}\leq\eta_{c}\tilde{D}_{p_{r},C_{p}},\label{pointtoline}
\end{gather}

\noindent with $\eta_{c}\in(0,\infty)$. 

Similarly, to maintain the needle tip $\myvec t_{n} \in\mathbb{H}_p$ \emph{inside} the guiding cylinder $\myvec C_{g}=(\myvec c_{g}, R_{g})$ in step 3 to prevent damaging the nearby tissue around the target, with $\myvec c_{g}= \myvec l_{g} + \dual\myvec m_{g}$ and $\myvec m_{g}=\myvec p_{p}\times\myvec l_{g}$, when performing the needle puncture, the following inequality was used

\begin{equation}
\dot{\tilde{D}}_{t,C_{g}}+\eta_{g}\tilde{D}_{t,C_{g}}\leq0\implies\frac{\partial D_{t,c_{g}}}{\partial\myvec q}\dot{\myvec q}\leq-\eta_{g}\tilde{D}_{t,C_{g}},\label{point-to-guideline}
\end{equation} 

\noindent with $\eta_{g}\in(0,\infty)$ and $\tilde{D}_{t,C_{g}}\triangleq D_{t,c_{g}}\left(\myvec t_{n}(\myvec q),\myvec c_{g}\right)-R_{g}^{2}$. 

To enforce the needle direction being inside the (infinite) cone $\myvec C_{o}=(\myvec c_{o},\theta_{o})$ surrounding the puncture guide line, where $\myvec c_{o}\in\mathcal{H}_p$ is the cone axis collinear with the puncture guide line (i.e., $\myvec c_{o}=\myvec c_{g}$) and $\theta_{o}\in(0,\pi/2)$ is the cone half-angle, a VFI is applied using a continuous bijective function named line-static-line-angle distance \cite{quir2019}. Denote the Plücker line $\myvec l_{\text{needle}} = \myvec l_{n} + \dual\myvec m_{n}$ collinear with the needle, where $\myvec l_{n}\triangleq\myvec l_{n}(\myvec q)\in\mathbb{H}_p$ is the needle direction and $\myvec m_{n}=\myvec t_{n}\times\myvec l_{n}$. By defining $\tilde{f}_{l_{n},c_{o}}\left(\phi_{l_{n},c_{o}}\right)\triangleq f_{l_{n},c_{o}}\left(\phi_{l_{n},c_{o}}\right)-f_{l_{n},c_{o}}\left(\theta_{o}\right)$, where $f_{l_{n},c_{o}}:\left[0,\pi\right]\rightarrow\left[0,4\right]$ with $f_{l_{n},c_{o}}\left(\phi_{l_{n},c_{o}}\right)=2-2\cos\phi_{l_{n},c_{o}}$, and $\phi_{l_{n},c_{o}}\triangleq\phi_{l_{n},c_{o}}(\myvec q)$ is the angle between $\myvec l_{n}$ and $\myvec c_{o}$ \cite{quir2019}, the inequality ensuring a maximum allowable needle misalignment was given by

\begin{equation}
\dot{\tilde{f}}_{l_{n},c_{o}}+\eta_{o}\tilde{f}_{l_{n},c_{o}}\leq0\implies\frac{\partial f_{l_{n},c_{o}}}{\partial\myvec q}\dot{\myvec q}\leq-\eta_{o}\tilde{f}_{l_{n},c_{o}}.\label{line-to-line}
\end{equation}

As $\tilde{f}_{l_{n},c_{o}}\left(\phi_{l_{n},c_{o}}\right)$ is a continuous bijective function, controlling the distance function \eqref{line-to-line} is equivalent to controlling the angle $\phi_{l_{n},c_{o}}\in\mathbb[0,\pi]$.

To prevent the needle tip from being inserted excessively and damaging the posterior tracheal wall during the puncture, a plane was defined $\myvec P_{p} = (\myvec l_{g},d_{P})$ passing through the target puncture position $\myvec p_{p}$ at the middle of the trachea cross-section with normal given by $\myvec l_{g} \in\mathbb{H}_p$, such that the distance to the reference frame is given by $d_{P}=\myvec p_{p}\cdot\myvec l_{g}$. To prevent the needle tip $\myvec t_{n}$ from trespassing $\myvec P_{p}$, the signed Euclidean distance between the needle tip and the plane was defined, $d_{t_{n}, P}\triangleq d_{t_{n}, P}\left(\myvec t_{n}(\myvec q),\myvec P_{p}\right)$, and the following inequality was used

\begin{equation}
\dot{d}_{t_{n},P}+\eta_{P}d_{t_{n},P}\geq0\implies-\frac{\partial d_{t_{n},P}}{\partial\myvec q}\dot{\myvec q}\leq\eta_{P}d_{t_{n},P}.\label{point-to-plane}
\end{equation}

Lastly, the constraint matrix $\myvec B$ and vector $\myvec b$ in \eqref{quadratic} were defined as

\begin{align*}
\mymatrix B & \!=\!\begin{bmatrix}-\frac{\partial D_{p,c_{p}}}{\partial\myvec q}^{T} & \!\frac{\partial D_{t,c_{g}}}{\partial\myvec q}^{T} & \!\frac{\partial f_{l_{n},c_{o}}}{\partial\myvec q}^{T} & \!-\frac{\partial d_{t_{n},P}}{\partial\myvec q}^{T}\! & \mymatrix W(\myvec q)^{T}\end{bmatrix}^{T}\\
\myvec b & \!=\!\begin{bmatrix}\begin{array}{cc}
\eta_{c}\tilde{D}_{p_{r},C_{p}} & \negmedspace-\eta_{g}\tilde{D}_{t,C_{g}}\end{array} & \negthickspace-\eta_{o}\tilde{f}_{l_{n},c_{o}} & \negmedspace\eta_{P}d_{t_{n},P} & \negmedspace\myvec w(\myvec q)\end{bmatrix}^{T},
\label{Bb}
\end{align*}

\noindent where $\mymatrix W(\myvec q)\in{\mathbb{R}^{4n\times n}}$ and $\myvec w(\myvec q)\in{\mathbb{R}^{4n}}$, whose explicit expressions can be found in \cite{mari2022}, represent $4n$ linear constraints in the control inputs to enforce lower and upper bounds in the joint velocities and joint angles.

\subsection{The robotic puncture procedure\label{The-Robotic-Puncture}}

In step 1, the robot aligns the needle with the puncture guide line $\myvec l_{g}$, in which the control objective is only the needle rotation $\myvec r_n \in\mathbb{H}_p$ to ease the visual inspection of the adaptation outcome, as the needle alignment is easy to observe. In step 2, once the needle is aligned, the control objective becomes the needle tip pose $\mydual x_n \in \mathcal{H}$ with its rotation fixed, which is a unit dual quaternion with $\mydual x_n = \myvec r_n + \frac{1}{2}\dual\myvec t_n\myvec r_n$. The needle tip is then moved to the puncture start position $\myvec p_{s}$ (see Fig.~\ref{simulation}) located above $\myvec p_{p}$ with distance $d_{p}$ along $\myvec l_{g}$. To ensure the initial error due to parametric uncertainty is reasonably small in Step 2, the robot waits for the estimated needle tip to converge to the measured one before performing the puncture in Step 3. More specifically, we use \eqref{adaptation} until the estimated needle direction and its tip converge to a sufficiently close neighbourhood around the measured values. In our implementation, we defined the threshold distance of $1.5$ mm and $0.5^{\circ}$, similar to the precision of the electromagnetic sensors we used.

In step 3, when the robot is performing the needle puncture, the needle tip position is controlled instead of its pose to release DoFs, but the needle is constrained by a cone so that its maximum angle with respect to the puncture guide line is limited. The cylinder constraint covering the patient in steps 1 and 2 is deactivated to avoid violating the VFI. Trajectory interpolation is applied by defining a quintic polynomial to define both start and end velocities and accelerations as 0 for the needle tip \cite{spon2020}. There are three major reasons for not including orientation data in the trajectory planning. First, when the needle is not penetrating the tissue, its axis direction can deviate from the puncture guide line without affecting the procedure.  Furthermore, considering the short trajectory length of the tissue penetration  ($\leq$ 50 mm), the needle rotation during this stage is negligible. Second, the line-to-line angle between the needle and the puncture guide line is further constrained to ensure it is within acceptable limits. Third, the sensor accuracy varies depending on its distance from the transmitter. Although both the needle sensor and the bronchoscope sensor return the same vector,   they are not fully aligned due to different distances to the transmitter.

\begin{figure*}[ht]
\centering
\subfloat[Time evolution of the distance to destination.]{
\includegraphics[width=0.32\textwidth]{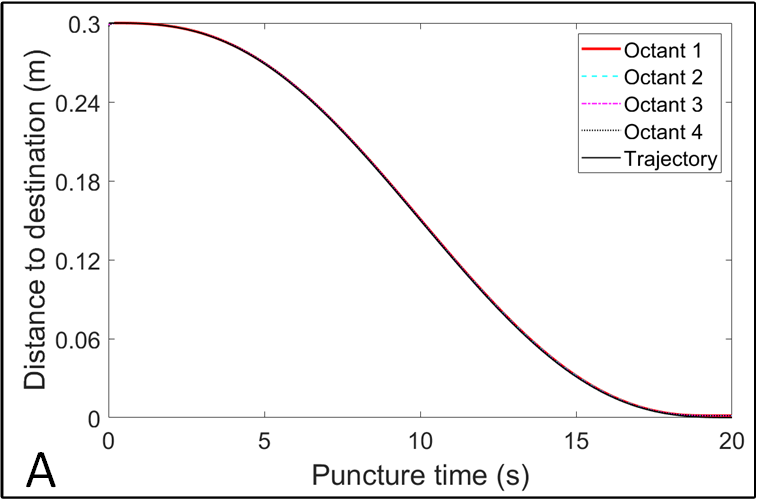}
}\subfloat[Distance from midline over time.]{
\includegraphics[width=0.32\textwidth]{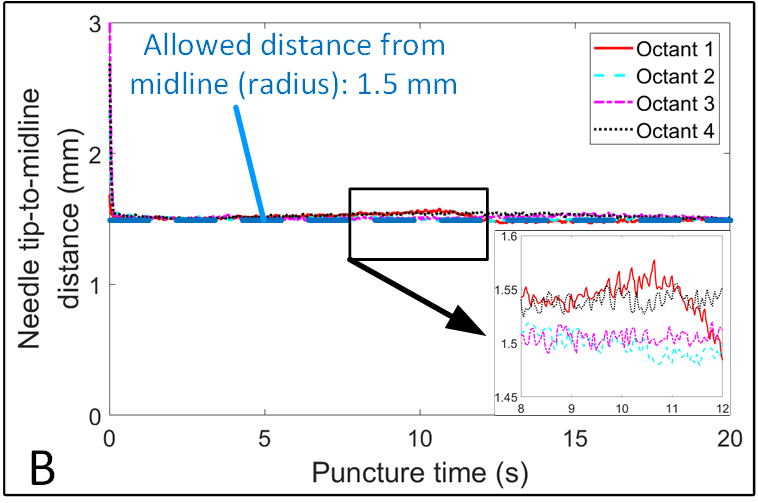}
}\subfloat[Needle's angle with respect to the guide line over time.]{
\includegraphics[width=0.32\textwidth]{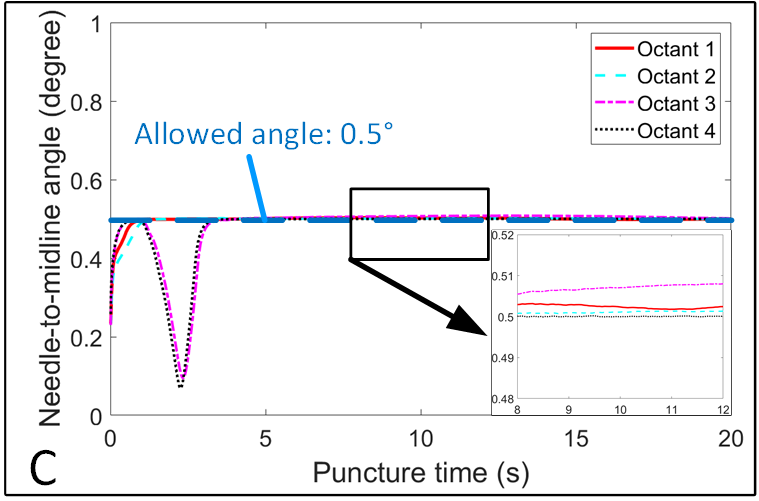}
}

\caption{Simulation results of the needle puncture. (A) The nominal quintic trajectory (\emph{black curve}) prescribes a smooth approach closely executed in the closed-loop system; (B) at the start of the robot motion, the error is larger than the threshold, but it is quickly reduced to satisfy the constraints. (C) The small violations ($<0.01^{\circ}$) are due to noisy measurements. \label{Simulationresult1}}
\end{figure*}

\begin{figure*}[ht]
\centering
\subfloat[Time response of the estimated robot base pose error under different
sets of initial uncertain kinematic parameters.]{
\includegraphics[width=0.48\textwidth]{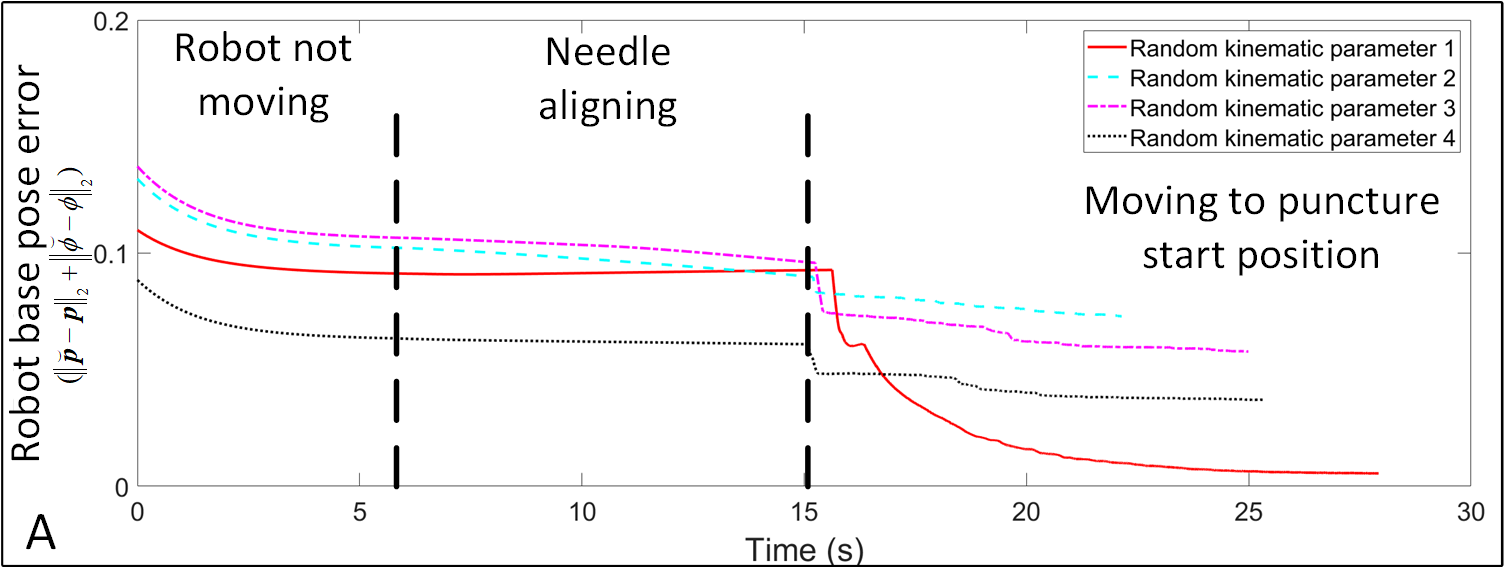}
}\subfloat[Time response of the estimated needle tip pose error.]{
\includegraphics[width=0.48\textwidth]{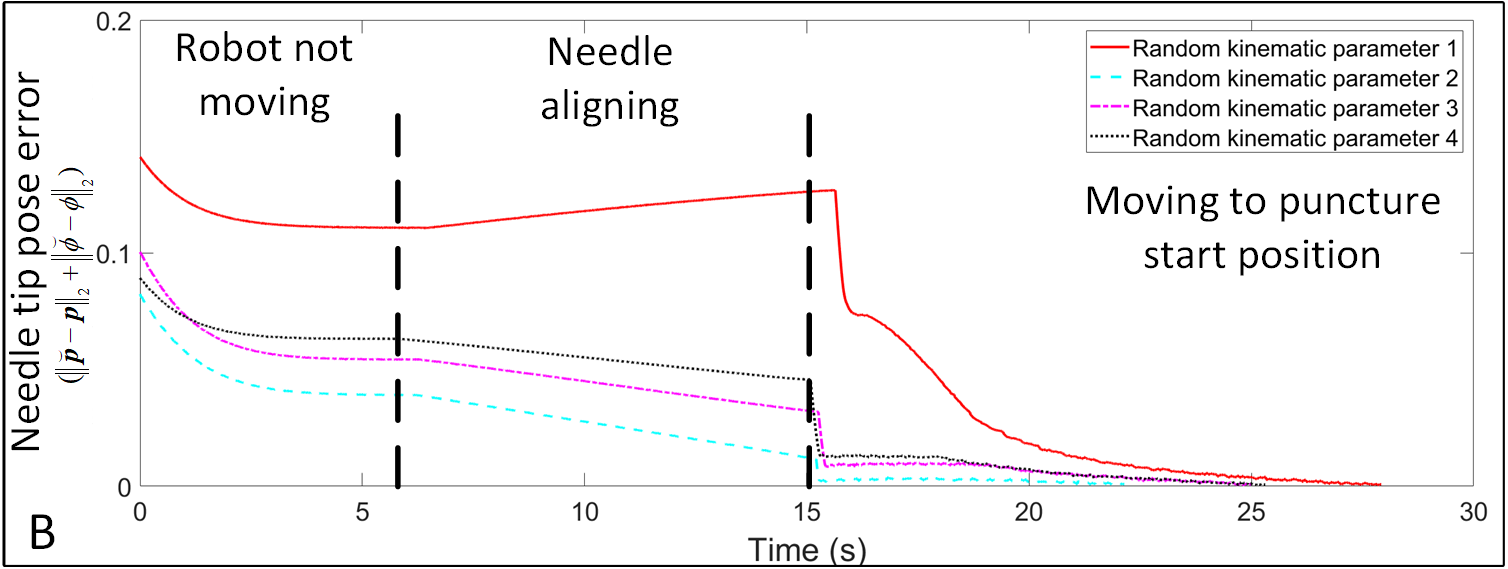}
}

\subfloat[Time response of the real distance to the puncture start position.]{
\includegraphics[width=0.48\textwidth]{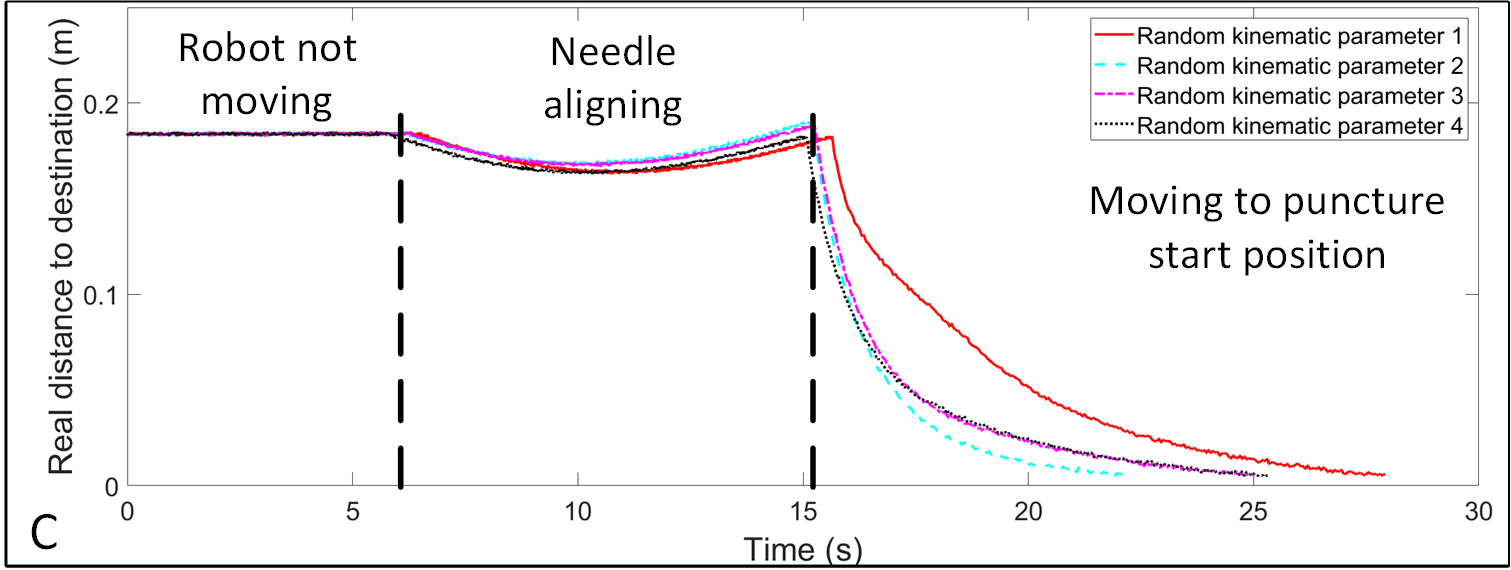}
}\subfloat[Time response of the real needle-to-midline angle. From 15 seconds onward, the maximum allowed angle between the guide line and the needle is $0.5^{\circ}$.]{
\includegraphics[width=0.48\textwidth]{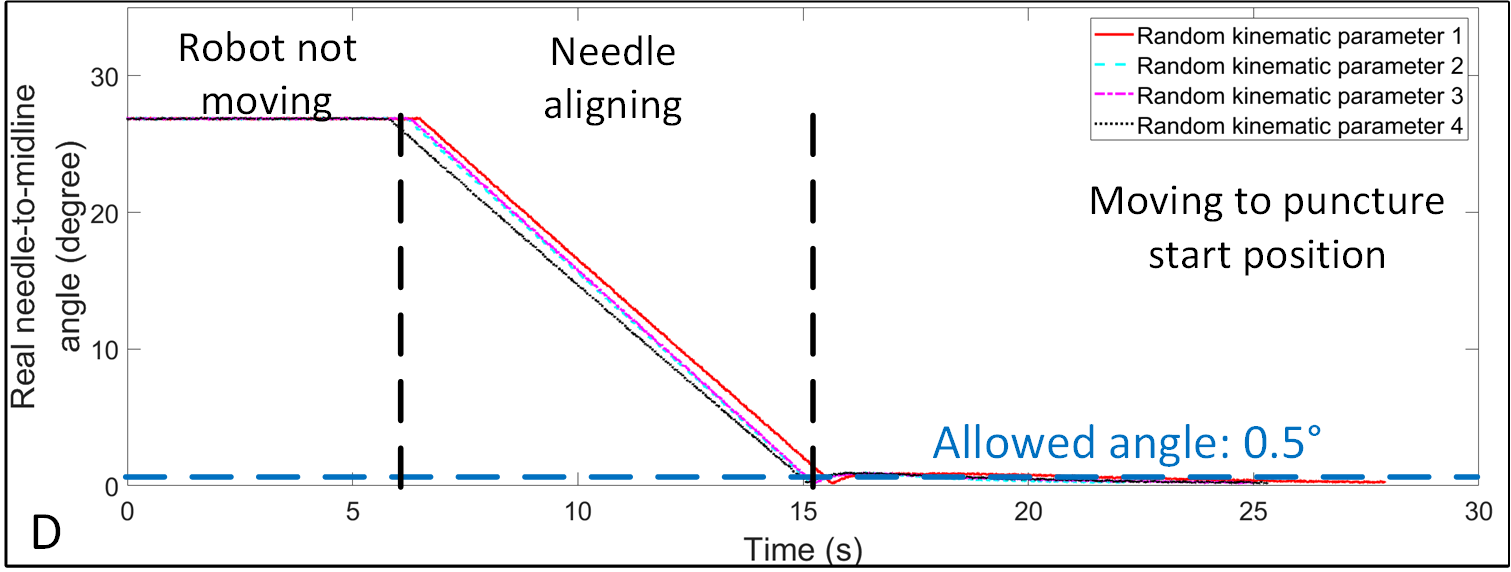}
}

\caption{Simulation results of the adaptation law. \label{simulation-plots}}
\end{figure*}

\section{Validation in simulation}
\subsection{Methodology of simulation}

Simulations were conducted using CoppeliaSim (Coppelia Robotics, Ltd., Switzerland) to evaluate the implementation of the adaptive constrained controller given by \eqref{quadratic}, with \eqref{adaptation}.\footnote{https://github.com/mmmarinho/tro2022\_adaptivecontrol}. The target puncture positions were randomly generated, which used a uniform distribution within a $10\times10\times10$ mm cube to simulate real sensor measurements made inside the trachea. The puncture guide line vectors were generated from the four octants above the plane (shown in Fig.~\ref{simulation}) by randomly generating the $xyz$ components to simulate different patient anatomy. The puncture start position is $d_{p}=300$ mm from the insertion point along the puncture guide line, which was considered sufficiently far from the patient for collision avoidance but within the transmitter range ($\pm450$ mm in any direction). An ideal Franka Emika Panda robot (Franka Robotics GmbH, Germany) with nominal geometric parameters was used. To simulate uncertain parameters, they were randomly generated around the nominal ones, using a uniform distribution, within $\pm200$ mm for the base position and $\pm10{}^{\circ}$ for the base rotation angles with respect to the transmitter frame, $\pm10$ mm for the needle tip position and $\pm5{}^{\circ}$ for the needle rotation angles with respect to the robot end-effector frame. The trajectory for step 3, described in Section \ref{Sensor-Guided-Constrained-Adapti}, was generated using a duration $T=20\:\unit{s}$. The cylinder preventing collisions with the patient in steps 1 and 2 has a radius $R_{p}=200$ mm, considering the shoulder widths of adults to be around 400 mm. In step 3, the maximum angle between the needle and the guiding cylinder was defined as $\theta_{o}=0.5^{\circ}$, whereas the guiding cylinder radius was defined as $R_{g}=1.5$ mm. All constraints were defined with respect to the transmitter frame. Three-dimensional Gaussian noise with zero mean and $1$ mm variance was added to the nominal needle tip and target puncture position, retrieved directly from CoppeliaSim, to simulate the noisy data measured by the real sensor. Gaussian noise with zero mean and $0.2^\circ$ variance was added to orientation measurements of both the simulated needle and bronchoscope sensors.

\subsection{Results and discussion of simulation}

\begin{figure*}[ht]
\centerline{\includegraphics[width=1\textwidth]{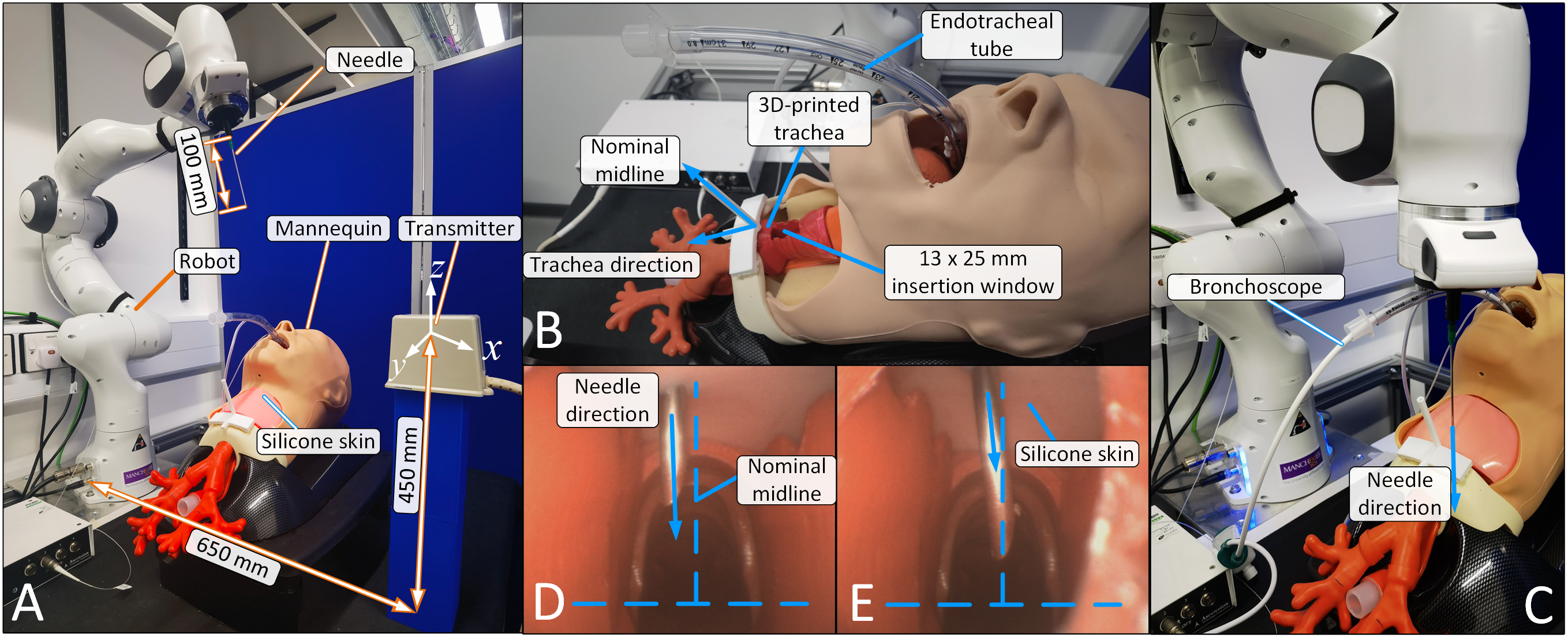}}
\caption{(A) Experimental setup of the robotic PDT needle puncture. (B) A mannequin was used for experiments. (C) Example of a needle puncture. (D-E) Bronchoscope images of two needle punctures performed on the silicone skin.}
\label{Puncture experiments}
\end{figure*}

Fig.~\ref{Simulationresult1}(A-C) illustrates the position and angular errors during the needle puncture with examples from four octants. All simulation trials followed the designated trajectory and maintained the position and angular errors within the bounds defined by the VFIs. Since the measured needle tip pose is noisy, the measured pose errors may sometimes violate those bounds slightly, but no more than the magnitude of the noise ($1$ mm), as shown in Fig.~\ref{Simulationresult1}(B). During the puncture, the position error was initially larger than the threshold, but it was quickly reduced to satisfy the constraints. The simulation was performed 100 times, and the mean position error between the target puncture position and the measured real needle tip, retrieved from CoppeliaSim as ground-truth, was $1.55\pm0.40$ mm within the range $[0.64,2.34]$ mm The mean angle between the puncture guide line and the needle was $0.38\pm0.021^{\circ}$ within the range $[0.08^{\circ},0.71^{\circ}]$.

Fig.~\ref{simulation-plots}(A-D) illustrates the adaptation process using four randomly generated sets of initial uncertain geometric parameters, as done in \cite{mari2022}, associated with the placement of the transmitter with respect to the robot base and the needle attachment. The robot first remained still while only the adaptation law \eqref{adaptation} ran until the adapted parameters' trajectories achieved steady state (Fig.~\ref{simulation-plots}(A) and \ref{simulation-plots}(B)), meaning that a local optimum had been achieved for the estimated parameters. Then, the complete motion control law (i.e., \eqref{quadratic} with \ref{adaptation}) was run so that the robot aligned the estimated needle direction with the guiding cylinder (Fig.~\ref{simulation-plots}(D)), then moved the estimated needle tip to the puncture start position with an accuracy of $3$ mm, and finally waited for the adaptation of the kinematic parameters before making the puncture (Fig.~\ref{simulation-plots}(B)). The position errors were not extracted from needle pose data, as we determined the quality of adaptation by observing if the needle pose errors converged to zero. The exact duration of adaptation was not recorded. The simulation results confirmed that the adaptation law we used could adapt the uncertain kinematic parameters caused by the measurement uncertainties and minimise the associated needle positioning and orienting error before performing the puncture.

\section{Experimental validation}

Validation experiments were conducted on a mannequin to evaluate the outcome of robotic PDT needle punctures. The goal was to evaluate the accuracy of the needle puncture performed by a robotic manipulator running our adaptive constrained control law in terms of small needle position and angular errors.

\subsection{Methodology of experiments}

The experimental setup is illustrated in Fig.~\ref{Puncture experiments}(A-B). A 14-gauge needle ($100$ mm length and $2.5$ mm outer diameter) from a PDT kit (Cook Critical Care, Limerick, Ireland) was installed on a 7-DoF Franka Emika Panda robot (control rate of 1000 Hz). Two Model 130 electromagnetic sensors designed for medical applications (NDI, Ontario, Canada), measuring 6-DoF pose information at 100 Hz, were used to guide the puncture. The sensors have a tip diameter of $1.3$ mm and a wire diameter of $0.9$ mm. One sensor was attached inside the needle tip, and the needle was mounted on a 3D-printed flange connected to the robot distal. The needle movement was achieved by the robot arm without using an independent actuator. The other sensor was inserted through the bronchoscope channel (Ambu A/S, Ballerup, Denmark) until it reached the bronchoscope's tip. The electromagnetic transmitter was placed at $\left(650,0,450\right)$ mm with an accuracy of $\pm10$ mm with respect to the robot base frame to cover the entire needle movement with its performance range. It remains \textbf{fixed} throughout the whole experiment \textbf{across all trials}. The transmitter's frame axes were parallel with the robot base frame axes with an accuracy of $\pm5^{\circ}$ measured by a protractor. A 3D-printed trachea model with a $13$ mm inner diameter and a $13\times25$ mm rectangular needle insertion window created before the experiment was installed on the mannequin (TruCorp Ltd., Ireland). An $8$ mm inner diameter endotracheal tube (Mallinckrodt Pharmaceuticals, MO, USA) was inserted through the mannequin's mouth to simulate the setup in a real PDT. The trachea was covered by a $10$ mm thick silicone artificial skin (made by a suture pad for clinical training) to simulate the pretracheal tissue of a non-obese adult. The constraints for robot motion are the same as in the simulation.

A calibration was performed before each trial to identify the rotation between the bronchoscope sensor frame and the bronchoscope image frame according to the following steps. First, the bronchoscope tip is placed on a horizontal plane (such as a table). Its position is held when the horizontal plane is also horizontal in the image, indicating that the vertical edge of the image is vertical in 3D space. The orientation of the bronchoscope sensor is recorded at the same time, indicating the bronchoscope sensor frame. Considering that the bronchoscope sensor is parallel to the bronchoscope tip axis, the vertical edge of the image can be quantitatively obtained by rotating the bronchoscope sensor frame around the bronchoscope tip axis.

Each trial is conducted in the following three steps:

\begin{itemize}
\item \textbf{T1}: An electromagnetic sensor is used to manually obtain the nominal midline and trachea direction vectors (see Fig.~\ref{Puncture experiments}(B)). This is to calculate the nominal midline plane. Calibration of the bronchoscope sensor was performed before each trial.
\item \textbf{T2}: The bronchoscope is inserted manually through the endotracheal tube until it reaches the tube's tip. The puncture guide line is obtained by manually aligning the nominal midline vertically as seen in the bronchoscope image. The trachea direction was obtained together with the puncture guide line. Then, the target puncture position is obtained by placing the bronchoscope sensor between two tracheal rings. 
\item \textbf{T3}: The robot is then activated when the operator is outside the workspace. The robot first aligns the needle with the puncture guide line obtained in T2 and moves to the puncture start position at $300$ mm above the target puncture position along the puncture guide line. Finally, the robot moves straight along the puncture guide line according to a quintic trajectory and punctures through the silicone pad in 25 seconds.
\end{itemize}

In the experiments, the mannequin was placed arbitrarily at four positions and orientations in the robot workspace to simulate different patient anatomies. The mannequin positions were in the range of $[150, 250]$ mm, $[-150, 0]$ mm and $[-100, 0]$ mm along the transmitter's $\boldsymbol{xyz}$ axes (see Fig.~\ref{Puncture experiments}(A)), respectively. The mannequin orientations varied in the rotation around the $\boldsymbol{z}$-axis of the transmitter within the range of $\pm30^\circ$. For each mannequin position, 100 needle punctures were performed. The silicone skin was replaced every 25 trials. Considering clinical expertise may affect the measurement accuracy, all manual tasks in each trial were performed by a non-clinical operator so that errors associated with the robot control and hardware can be better identified and analysed.

\begin{figure}[ht]
\centerline{\includegraphics[width=1\columnwidth]{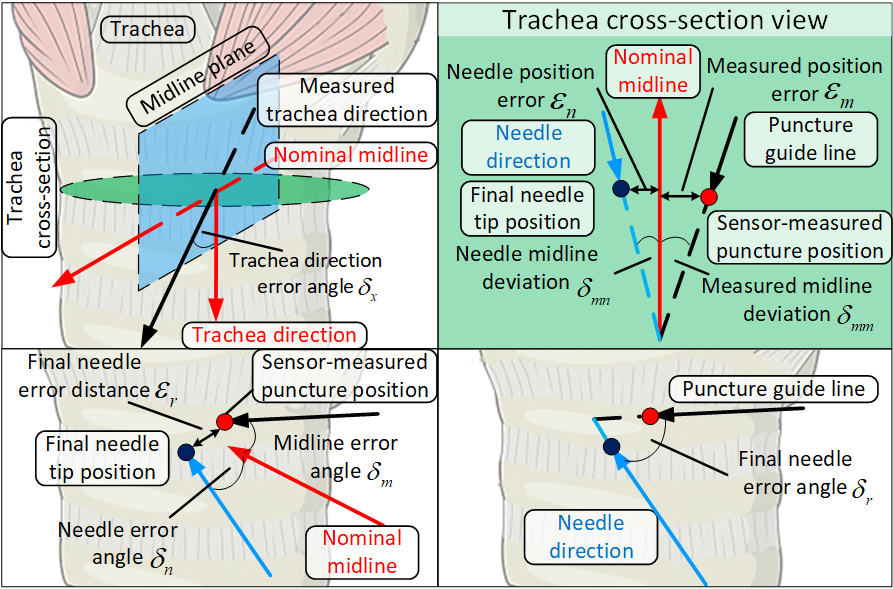}}
\caption{Illustration of variables collected during the experiments (seen from anterior to posterior tracheal wall). Variables $\delta_{mn}$, $\delta_{n}$ and $\varepsilon_{n}$ determine the quality of puncture. Variables $\delta_{mm}$, $\delta_{m}$ and $\varepsilon_{m}$ determine the quality of bronchoscope sensor measurements. Variables $\delta_{r}$ and $\varepsilon_{r}$ evaluate the VFI constraint violations.}
\label{Puncture illustration}
\end{figure}

All position and orientation data were measured by the bronchoscope sensor and were averaged from a 1-second recording at 100 Hz. Data were analysed and visualised using SPSS 24 (IBM, New York, USA) to evaluate the accuracy of the needle puncture and sensor measurements. The position and angular variables collected during the experiments are illustrated in Fig.~\ref{Puncture illustration}. All values were computed from sensor measurements after the needle finally completed the puncture. The data distributions were examined with Shapiro-Wilk tests of normality. Summary statistics were reported as mean $\pm$ standard deviation (SD) for normally distributed data or median (interquartile range--IQR) otherwise.

\begin{figure*}[ht]
\centering
\subfloat[Translation errors for four mannequin positions.]{
\includegraphics[width=0.3\textwidth]{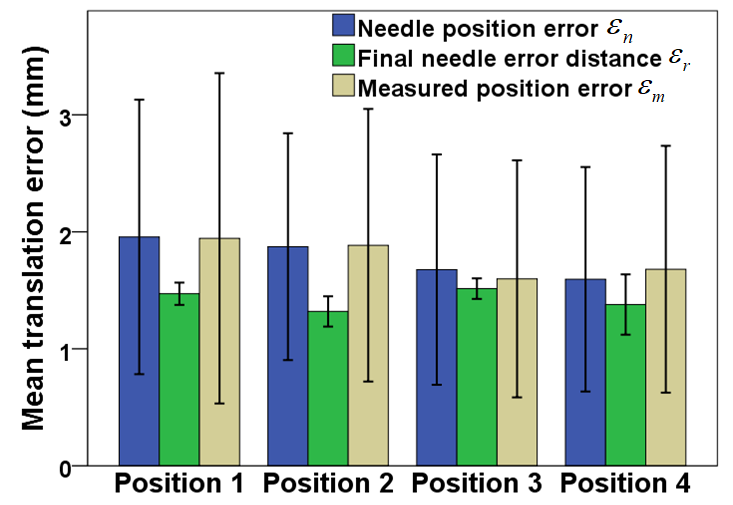}
}\hspace{0.02\textwidth}\subfloat[Angular errors for four mannequin positions.]{
\includegraphics[width=0.3\textwidth]{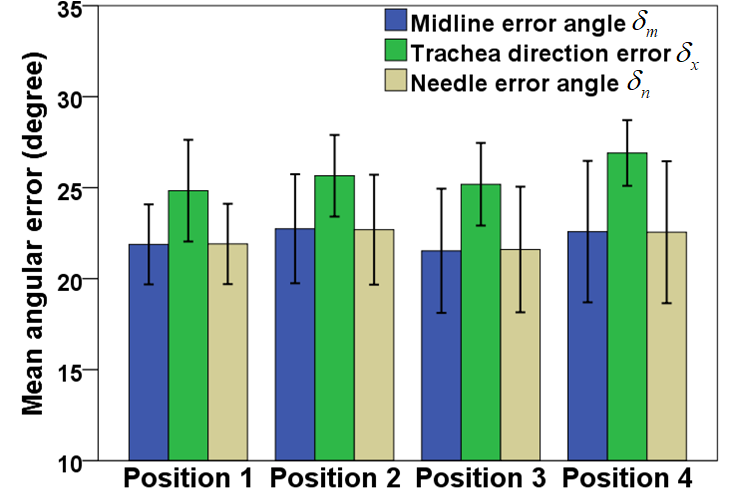}
}\hspace{0.02\textwidth}\subfloat[Midline deviations and final needle angles for four mannequin positions.]{
\includegraphics[width=0.3\textwidth]{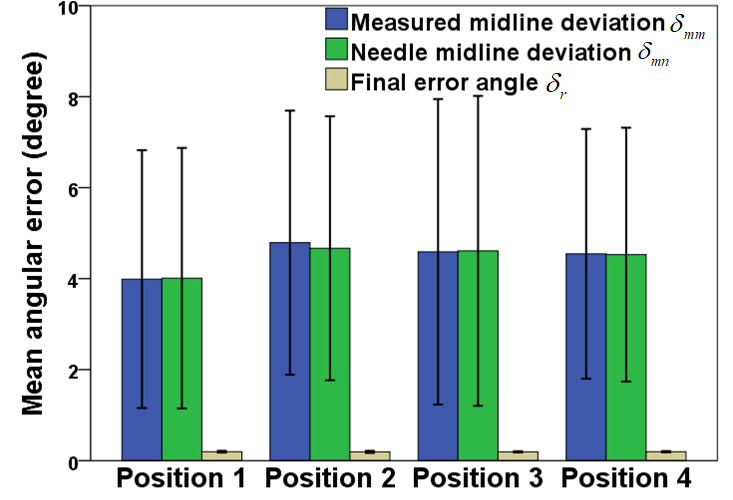}
}
\caption{Experiment results of the robotic puncture. \label{Puncture experiment2}}
\end{figure*}

\begin{table*}[ht]
\caption{Results of robotic PDT needle puncture experiments$^{1,2}$}
\label{Puncture results1}
\begin{threeparttable}
\begin{tabular}{>{\raggedright}m{0.36\textwidth}>{\raggedleft}p{0.1\textwidth}>{\raggedleft}p{0.1\textwidth}>{\raggedleft}p{0.1\textwidth}>{\raggedleft}p{0.1\textwidth}>{\raggedleft}p{0.1\textwidth}}
\toprule 
 & {\textbf{Mannequin position 1}} & {\textbf{Mannequin position 2}} & {\textbf{Mannequin position 3}} & {\textbf{Mannequin position 4}} & {\textbf{All mannequin positions}}\tabularnewline
\midrule
{\textbf{Number of trials}} & { 100} & { 100} & { 100} & { 100} & { 400}\tabularnewline

{\textbf{Median (IQR) of needle tip position error $\varepsilon_{n}$ [\unit{\milli\meter}]}} & { 2.1 (2.3)} & { 1.7 (1.8)} & { 1.7 (1.7)} & { 1.6 (1.5)} & { 1.7 (1.9)}\tabularnewline

{\textbf{Median (IQR) of final needle error distance $\varepsilon_{r}$ [\unit{\milli\meter}]}} & { 1.5 (0.1)} & { 1.3 (0.2)} & { 1.5 (0.1)} & { 1.4 (0.3)} & { 1.4 (0.2)}\tabularnewline

{\textbf{Median (IQR) of needle error angle $\delta_{n}$ [\unit{\degree}]}} & { 21.60 (2.86)} & { 22.69 (5.14)} & { 21.60 (4.79)} & { 23.43 (4.42)} & { 21.91 (4.60)}\tabularnewline

{\textbf{Median (IQR) of final needle error angle $\delta_{r}$ [\unit{\degree}]}} & { 0.20 (0.01)} & { 0.20 (0.01)} & { 0.20 (0.02)} & { 0.20 (0.02)} & { 0.20 (0.02)}\tabularnewline

{\textbf{Median (IQR) of needle midline deviation $\delta_{mn}$ [\unit{\degree}]}} & { 3.33 (3.74)} & { 4.55 (4.66)} & { 4.61 (4.71)} & { 4.60 (5.31)} & { 4.13 (4.55)}\tabularnewline
\bottomrule
\end{tabular}
\begin{tablenotes}\footnotesize
\item[1] Refer to Fig.~\ref{Puncture illustration} for definitions of the variables.
\item[2] Refer to Table \ref{Puncture resultsA} in Appendix for more results.
\end{tablenotes}
\end{threeparttable}
\end{table*}

Variables to evaluate the quality of puncture are (1) needle midline deviation $\delta_{mn}$ (line-to-line angle between the needle and nominal midline, projected onto the trachea cross-section plane), (2) needle midline error angle $\delta_{n}$ (line-to-line angle between the needle and nominal midline), and (3) position error of the needle tip $\varepsilon_{n}$ (point-to-plane distance between the needle tip and midline plane). In clinical practice, the midline deviation is required to be smaller than $30^{\circ}$ \cite{ruda2014,kuma2022}, so $\delta_{mn}$ must fall within that range to meet the clinical requirements. Regarding the lower limit of the inner diameter of a normal adult trachea to be 10--13 mm \cite{brea1984}, the puncture position error $\varepsilon_{n}$ must be within $\pm5$ mm, assuming that an ideal needle insertion must be at the center of a trachea with a 10 mm inner diameter. Otherwise, the needle puncture might miss the trachea. 

Variables to evaluate if the VFI constraints are violated are (1) final needle error angle $\delta_{r}$ (line-to-line angle between the needle and the puncture guide line), and (2) final needle error distance $\varepsilon_{r}$ (distance between the needle tip and the target puncture position). The value $\varepsilon_{r}$ indicates how close to the guiding cylinder's boundaries the needle tip is, whereas $\delta_{r}$ indicates the needle angle with respect to the puncture guide line. In the experiments, the upper limits for $\varepsilon_{r}$ and $\delta_{r}$ are $1.5$ mm and $0.5^{\circ}$, respectively. 

Clinically, an ideal puncture should be placed exactly on the midline plane, shown in Fig.~\ref{Puncture illustration}, following the nominal midline. Values to evaluate the quality of bronchoscope sensor measurements are (1) midline deviation measurement error $\delta_{mm}$ (line-to-line angle between the puncture guideline, projected onto the trachea cross-section plane, and the nominal midline), (2) midline error angle $\delta_{m}$ (line-to-line angle between the nominal midline and puncture guide line), (3) measured position error $\varepsilon_{m}$ (point-to-plane distance between the target puncture position measured by the bronchoscope sensor and the midline plane in Fig.~\ref{Puncture illustration}), and (4) trachea direction error angle $\delta_{x}$ (line-to-line angle between nominal and measured trachea direction). The values $\delta_{mm}$ and $\delta_{m}$ indicate how close the puncture guide line obtained by the sensor and sent to the robot is to the nominal midline. To identify how the bronchoscope sensor measurements affect the puncture accuracy, Pearson (for normally distributed data) or Spearman (for non-normal data) correlation tests were conducted for $\delta_{m}$, $\delta_{n}$ and $\varepsilon_{m}$, $\varepsilon_{n}$ to identify the linear correlation between these values. If the linear correlation is identified, then we can conclude that improving the accuracy of the bronchoscope sensor measurement will lead to more accurate robotic PDT needle puncture, which justifies improving the technique of bronchoscope sensor measurement in future work. 

\section{Results}

In total, 400 robotic PDT needle punctures were performed for four mannequin configurations, each of which was done after a preprocedural measurement of nominal vectors, target puncture position, and guide line. Nineteen of the total of 419 trials (4.53\%) failed, and the robot stopped partway during the procedure due to the safety shutdown triggered by the loss of communication with the control laptop. The normality of all results was checked by Shapiro-Wilk tests, and all results were non-normally distributed. All trials were performed within three minutes (from the beginning of bronchoscope insertion to the end of puncture). Key results are illustrated in Table~\ref{Puncture results1}. All other results are illustrated in Fig.~\ref{Puncture experiment2} and Table~\ref{Puncture resultsA} in the Appendix.

Spearman correlation tests were performed between needle errors and measurement errors to determine if they are linearly correlated. Monotonic relationships were confirmed between $\varepsilon_{m}$, $\varepsilon_{n}$ and between $\delta_{m}$, $\delta_{n}$, which satisfied the assumption of the Spearman test. The Spearman correlation coefficient $r$ between $\varepsilon_{m}$ and $\varepsilon_{n}$ is 0.909 ($P<0.001$), indicating the strong positive and linear relationship between $\varepsilon_{m}$ and $\varepsilon_{n}$ since $r$ is close to 1 \cite{myer2014}. The same linear relationship was obtained between $\delta_{m}$ and $\delta_{n}$ with $r$ being 0.999 ($P<0.001$). The correlation between $\varepsilon_{m}$, $\varepsilon_{n}$ and between $\delta_{m}$, $\delta_{n}$ are illustrated in Fig.~\ref{correlate} in the Appendix.

\section{Discussion}

In clinical practice, the midline deviation is required to be smaller than $30^{\circ}$ \cite{ruda2014,kuma2022} and the puncture position error must be within $\pm5$ mm in order not to miss the trachea \cite{brea1984}. The simulation results have validated the feasibility of the adaptive constrained controller for PDT, and the experiments have quantitatively determined the accuracy of robotic needle puncture in PDT using a mannequin. Potential improvements of robotic PDT needle puncture compared with current techniques are: (1) reducing the position and angular errors of the needle puncture, (2) reducing the difficulty of the PDT procedure.

The experiment results showed that all values of the position error $\varepsilon_{n}$ between the inserted needle tip position and the nominal midline plane are smaller than the allowable threshold (5 mm) and satisfy the accuracy requirements we developed in previous work \cite{tang2023}. 95.47\% of the needle punctures were successfully performed through the insertion window illustrated in Fig.~\ref{Puncture experiments}(B). The position errors are mainly attributed to imprecise bronchoscope sensor placements inside the trachea to obtain the reference target point. Since the bronchoscope image cannot indicate the exact location of the sensor in 3D space, as it does not provide binocular cues, such as depth, it is difficult to move the sensor onto the midline plane. In our experiments, one potential reason for having inaccurate bronchoscope sensor measurements is the lack of clinical experience, as the operator was not a clinician. The measured position error $\varepsilon_{m}$ (point-to-plane distance between midline plane and the target puncture position measured by the bronchoscope sensor) reaches up to $4.1$ mm and sometimes larger than $\varepsilon_{n}$. This is because the cylinder constraint in step 3 makes the final needle tip location be within 1.5 mm radial distance to the measured target puncture position (i.e., $\varepsilon_{m} \leq 1.5$ mm). However, because of a lack of depth cues when measuring the target puncture position, the target puncture might be a bit far from the midline plane, increasing the value of $\varepsilon_{m}$. Because $\varepsilon_{r}$, $\varepsilon_{m}$, and $\varepsilon_{n}$ are not independent, it might be the case that the final needle tip position is closer to the midline plane than the target puncture position, so that $\varepsilon_{n} < \varepsilon_{m}$. Conversely, for the same value of $\varepsilon_{r}$, if the final needle tip position is farther from the midline plane than the target puncture position is, then $\varepsilon_{n} > \varepsilon_{m}$.

Compared with another quantitative study \cite{haro2024} for manual PDT insertions on a mannequin, which used the same metrics as position error $\varepsilon_{r}$, our results achieved smaller position errors using the same mannequin and electromagnetic sensors (Median: 1.4 mm vs. 2.9 mm, IQR: 0.2 mm vs. 2.1 mm).

The results of using electromagnetic sensors to guide the robotic puncture meet the clinical requirements regarding midline deviation and position errors. The midline deviation of the needle $\delta_{mn}$ of all trials falls into the clinical requirements ($\pm30^{\circ}$) \cite{ruda2014,kuma2022}. Compared with the palpation-only technique ($35.00^{\circ}\pm5.00^{\circ}$) and real-time ultrasound assistance ($15.00^{\circ}\pm3.00^{\circ}$) \cite{ruda2014,kuma2022}, the use of electromagnetic sensors could potentially lead to more accurate acquisition of the puncture guide line due to smaller midline deviation measurements, as shown in our experiments. 

\begin{figure}[tbh]
\centerline{\includegraphics[width=1\columnwidth]{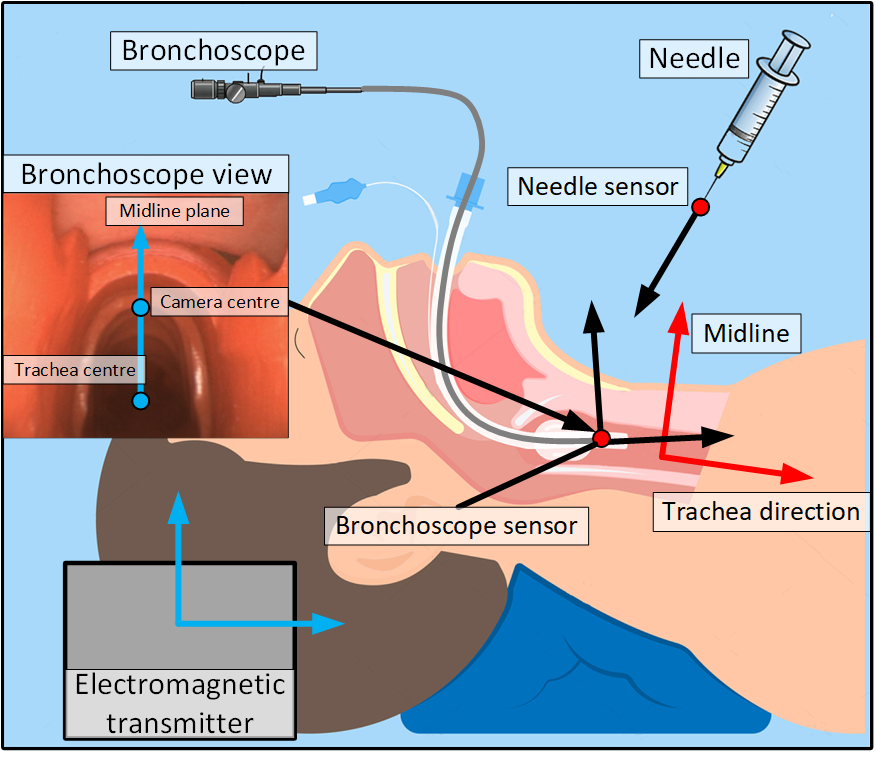}}
\caption{Illustration of the bronchoscope sensor measurement. The bronchoscope sensor may not be fully aligned with the nominal trachea direction, which will lead to angular errors in puncture guide line measurements.}
\label{Illustration2}
\end{figure}

The angle $\delta_{n}$ is 498.65\% larger than $\delta_{mn}$ and the final needle error angle $\delta_{r}$ is 95.5\% smaller than $\delta_{n}$. The values of $\delta_{r}$ are more precise than $\delta_{n}$ and $\delta_{mn}$ as seen from the smaller standard deviations. Therefore, the major angular error source is the bronchoscope sensor measurements due to inaccurate manual bronchoscope insertion. This is because the bronchoscope used in the experiments has only one bending DoF, making it difficult to align with both the nominal midline and trachea direction simultaneously. As illustrated in Fig.~\ref{Illustration2}, when obtaining the puncture guide line, the bronchoscope tip is not fully aligned with the trachea direction, having the median error angle $\delta_{x}$ of $25.48^{\circ}$. This is confirmed from the bronchoscope images (such as Fig.~\ref{Illustration2}) in which the trachea center is lower than the image/camera center, indicating the bronchoscope tip is pointing more upwards than the nominal trachea direction. The puncture guide line will also deviate since it is perpendicular to the sensor's axis $\boldsymbol{x}_{B}$. Although the measured midline deviation $\delta_{mm}$ might have a small error (such as when the $\boldsymbol{z}$-axis of the bronchoscope sensor $\boldsymbol{z}_{B}$ falls into the midline plane, the puncture guide line may become too cranial (close to the patient's head) or caudal (close to the patient's feet). Furthermore, the sensor may not perfectly fall into the midline plane, in which the camera center will be either left or right relative to the trachea midline seen from the bronchoscope images.

The feasibility of the constraints we implemented for robotic PDT needle puncture has been validated. Considering the geometric constraints we defined for the needle puncture, the error distance $\varepsilon_{r}$ is expected to be within $1.50$ mm and the error angle $\delta_{r}$ to be within $0.50^{\circ}$ to avoid violating the constraints. Experimental results have illustrated that the error angle $\delta_{r}$ reaches up to $0.36^{\circ}$ and has no violation. The error distance $\varepsilon_{r}$ occasionally goes above the threshold value and reaches up to $2.00$ mm. Reasons that contribute to this behaviour are: (1) the measuring values from the sensor have small errors up to $1$ mm due to its accuracy; (2) the accuracy of the sensor measuring is affected by factors such as sensor orientation and distance to the transmitter, which may change while moving the needle \cite{shen2008}; (3) the needle may bend slightly under the contact force during the insertion due to not being perpendicular to the silicone surface and its length. The associated angular error caused by deformation could be compensated for by reducing the needle length, but the tip offset is more difficult to eliminate due to the resisting force from the silicone. Using a more accurate sensor, increasing the needle strength, or reducing its length could reduce the position error of the robotic puncture.

Since the values of measured position error $\varepsilon_{m}$ and error $\varepsilon_{r}$ are close and may interfere with each other, we conducted Spearman correlation tests to verify the claim that reducing $\varepsilon_{m}$ will also reduce $\varepsilon_{r}$. Results have demonstrated a strong positive and linear relationship between $\varepsilon_{m}$ and $\varepsilon_{r}$ due to the coefficient $r$ close to 1 with P value $<$0.001 \cite{myer2014}. The test results between angular errors $\delta_{m}$ and $\delta_{n}$ also illustrate a strong positive and linear relationship. Therefore, bronchoscope sensor measurements and needle puncture errors are strongly correlated, and the claim is verified. Thus, we conclude that reducing the errors associated with the bronchoscope sensor measurements should improve the overall accuracy of the robotic puncture.

A major potential advantage of the proposed robotic system is that it releases the role of needle puncture from the human operator. In a manual PDT, at least two people are needed. One bronchoscopist holds the bronchoscope, and one operator performs the puncture. The main function of our system is to perform the puncture while the bronchoscopist manages the airway and observes the procedural outcome. The roles of another operator now are supervising the procedure and performing subsequent PDT dilation and tube insertion.

The major limitation is the dependence on the quality of sensor measurements. The accuracy of the puncture is correlated with the manual bronchoscope operation, as verified by the Spearman correlation tests. The position and angular errors of manual sensor placement inside the trachea could lead to large needle puncture errors since the robot relies on the sensor data to move. Nonetheless, routine clinical care involves direct visualization of the puncture site by bronchoscopy. Manipulating a target sensor to the optimum insertion point is technically challenging but well within the skillset of most experienced ICU clinicians who are involved in PDT. Furthermore, there are some limitations of the experiment setup, including using silicone to simulate neck tissue (difference in material properties), manual tasks performed by a non-clinical operator (difference in clinical expertise), unvaried trachea shape and dimension (lack of anatomical variance).

\begin{table*}[ht]
\caption{Results of robotic PDT needle puncture experiments (extended)$^{1}$}
\label{Puncture resultsA}
\begin{threeparttable}
\begin{tabular}{>{\raggedright}m{0.28\textwidth}>{\raggedleft}p{0.12\textwidth}>{\raggedleft}p{0.12\textwidth}>{\raggedleft}p{0.12\textwidth}>{\raggedleft}p{0.12\textwidth}>{\raggedleft}p{0.1\textwidth}}
\toprule
 & {\textbf{Mannequin position 1}} & {\textbf{Mannequin position 2}} & {\textbf{Mannequin position 3}} & {\textbf{Mannequin position 4}} & {\textbf{All mannequin positions}}\tabularnewline
\midrule

{\textbf{Mean $\pm$ SD of $\varepsilon_{m}$ [\unit{\milli\meter}]}} & { 1.9 $\pm$ 1.4} & { 1.9 $\pm$ 1.4} & { 1.6 $\pm$ 1.0} & { 1.7 $\pm$ 1.1} & { 1.8 $\pm$ 1.2}\tabularnewline

{\textbf{Mean $\pm$ SD of $\varepsilon_{n}$ [\unit{\milli\meter}]}} & { 2.0 $\pm$ 1.4} & { 1.9 $\pm$ 1.0} & { 1.7 $\pm$ 1.0} & { 1.6 $\pm$ 1.0} & { 1.8 $\pm$ 1.0}\tabularnewline

{\textbf{Mean $\pm$ SD of $\varepsilon_{r}$ [\unit{\milli\meter}]}} & { 1.5 $\pm$ 0.2} & { 1.3 $\pm$ 0.3} & { 1.5 $\pm$ 0.2} & { 1.4 $\pm$ 0.3} & { 1.4 $\pm$ 0.2}\tabularnewline

{\textbf{Median (IQR) of $\varepsilon_{m}$ [\unit{\milli\meter}]}} & { 2.0 (2.7)} & { 1.8 (1.7)} & { 1.5 (1.8)} & { 1.6 (1.8)} & { 1.7 (2.0)}\tabularnewline

\midrule

{\textbf{Mean $\pm$ SD of $\delta_{m}$ [\unit{\degree}]}} & { 21.88 $\pm$ 2.20} & { 22.74 $\pm$ 2.99} & { 21.53 $\pm$ 3.42} & { 22.58 $\pm$ 3.89} & { 22.18 $\pm$ 3.21}\tabularnewline

{\textbf{Mean $\pm$ SD of $\delta_{n}$ [\unit{\degree}]}} & { 21.91 $\pm$ 2.21} & { 22.69 $\pm$ 3.02} & { 21.60 $\pm$ 3.45} & { 22.55 $\pm$ 3.90} & { 22.19 $\pm$ 3.23}\tabularnewline

{\textbf{Mean $\pm$ SD of $\delta_{r}$ [\unit{\degree}]}} & { 0.20 $\pm$ 0.02} & { 0.19 $\pm$ 0.02} & { 0.19 $\pm$ 0.01} & { 0.20 $\pm$ 0.01} & { 0.20 $\pm$ 0.02}\tabularnewline

{\textbf{Mean $\pm$ SD of $\delta_{mm}$ [\unit{\degree}]}} & { 3.99 $\pm$ 2.83} & { 4.79 $\pm$ 2.90} & { 4.59 $\pm$ 3.36} & { 4.54 $\pm$ 2.75} & { 4.48 $\pm$ 2.97}\tabularnewline

{\textbf{Mean $\pm$ SD of $\delta_{mn}$ [\unit{\degree}]}} & { 4.01 $\pm$ 2.86} & { 4.67 $\pm$ 2.90} & { 4.61 $\pm$ 3.41} & { 4.53 $\pm$ 2.79} & { 4.45 $\pm$ 3.00}\tabularnewline

{\textbf{Mean $\pm$ SD of $\delta_{x}$ [\unit{\degree}]}} & { 24.84 $\pm$ 2.79} & { 25.65 $\pm$ 2.24} & { 25.19 $\pm$ 2.27} & { 26.91 $\pm$ 1.80} & { 25.65 $\pm$ 2.43}\tabularnewline

{\textbf{Median (IQR) of $\delta_{m}$ [\unit{\degree}]}} & { 21.56 (2.77)} & { 22.36 (5.12)} & { 20.39 (4.78)} & { 23.47 (4.49)} & { 21.86 (4.57)}\tabularnewline

{\textbf{Median (IQR) of $\delta_{mm}$ [\unit{\degree}]}} & { 3.49 (3.90)} & { 4.62 (4.72)} & { 4.21 (4.57)} & { 4.63 (4.93)} & { 4.20 (4.61)}\tabularnewline

{\textbf{Median (IQR) of $\delta_{x}$ [\unit{\degree}]}} & { 20.95 (4.30)} & { 25.99 (2.05)} & { 24.86 (3.30)} & { 26.81 (2.30)} & { 25.48 (3.80)}\tabularnewline
\bottomrule
\end{tabular}
\begin{tablenotes}\footnotesize
\item[1] Refer to Fig.~\ref{Puncture illustration} for definitions of the variables.
\end{tablenotes}
\end{threeparttable}
\end{table*}

Potential improvements of the robotic PDT technique could include the following three aspects:

\begin{itemize}
\item To decrease the dependency on direct visualization of the puncture site, future work will focus on techniques to quantitatively calculate rather than directly measure the target puncture position and direction based on mathematical modelling.
\item Assign more tasks for robotic PDT. Two critical steps for the PDT in current clinical practice are the needle puncture and dilation, while we automated the former using a robotic manipulator. Developing a technique to enable robotic dilation will further improve the PDT and reduce the procedural difficulty.
\item Currently, the robot control is based on kinematics, and considering that excessive force may harm the patient, haptic sensors could be implemented into the robot controller in further developments. 
\item Use animal tissues, such as a porcine trachea and skin, to better simulate the material properties of a human neck.
\end{itemize}

\section{Conclusions}

In this study, we proposed a new robotic system for PDT needle puncture to automate the most critical step for PDT. We implemented a constrained controller in which geometric constraints were defined to (1) cover the patient body to prevent robot-patient collision, and (2) constrain the needle motion to achieve an accurate PDT puncture with small position and angular errors. We also implemented an adaptive controller to dynamically adapt the uncertain kinematic parameters associated with the needle attachment and sensor transmitter placement to increase puncture accuracy, which enabled the accurate guidance of the robot movement using electromagnetic sensors. Both simulation and experimental validation were conducted to evaluate the performance of the robotic PDT needle puncture. Results have demonstrated high accuracy in terms of small needle position errors and midline deviations and high precision in terms of small standard deviations, suggesting the feasibility of using a robot to perform PDT puncture. Future work will focus on improving sensor measurements and developing a technique to enable robotic dilation to further automate the PDT.

\section*{Acknowledgement}

We thank the authors from \cite{zhan2022} for kindly providing the 3D model of the mannequin when making the demo video.

\section*{Appendix}

Table~\ref{Puncture resultsA} shows more data from the robotic PDT needle puncture experiments, complementing Table \ref{Puncture results1}.

\begin{figure}[h]
\centerline{\includegraphics[width=1\columnwidth]{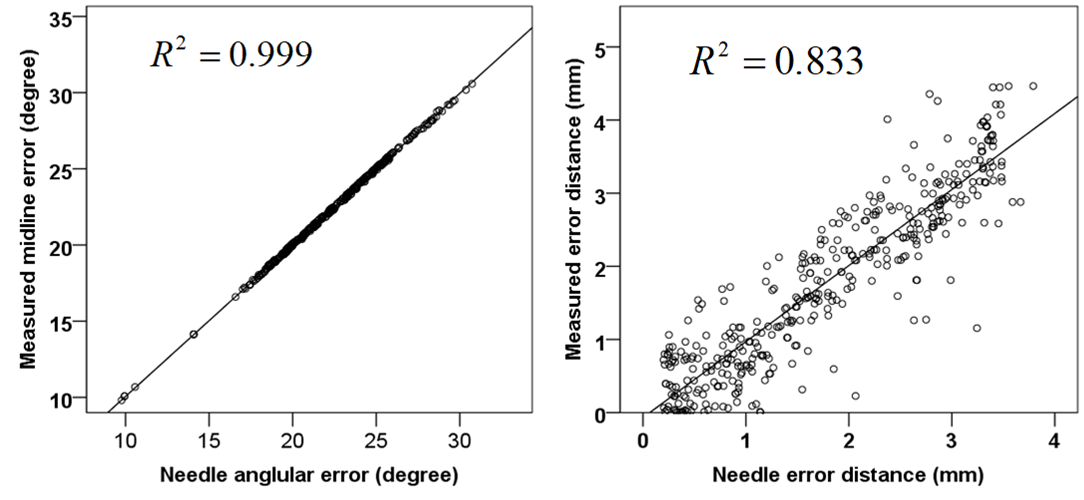}}
\caption{Correlation between $\delta_{m}$, $\delta_{n}$ (left) and $\varepsilon_{m}$, $\varepsilon_{n}$ (right).}
\label{correlate}
\end{figure}

Fig.~\ref{correlate} illustrates the correlation between $\delta_{m}$, $\delta_{n}$ and between $\varepsilon_{m}$, $\varepsilon_{n}$.

\AtNextBibliography{\footnotesize}
\printbibliography
\end{document}